%% file: main.tex
\documentclass[letterpaper, 10 pt, conference]{ieeeconf}

\IEEEoverridecommandlockouts
\usepackage{cite}
\usepackage{amsmath,amssymb,amsfonts}
\usepackage{algorithmic}
\usepackage{graphicx}
\usepackage{textcomp}
\usepackage{xcolor}
\usepackage{balance}
\usepackage{comment}
\usepackage[ruled,vlined]{algorithm2e}
\usepackage{caption}
\usepackage{array}
\usepackage{subcaption}
\usepackage{arydshln}
\usepackage{tensor}

\usepackage{hyperref}
\usepackage{float}
\usepackage{rotating}
\usepackage{nicematrix}  
\usepackage{mathtools}
\usepackage{pifont}
\usepackage{cuted}
\usepackage{dblfloatfix}

\definecolor{LightCyan}{rgb}{0.88,1,0.88}
\definecolor{linear_color}{RGB}{220,223,240}
\definecolor{gray_bbox_color}{RGB}{243,243,244}

\definecolor{rebuttal}{rgb}{0,0,1}

\def\eqref#1{Eq.~(\ref{#1})}

\newcommand{\ex}[1]{\texttt{#1}}

\usepackage{fancybox}
\usepackage{hhline}
\usepackage{bm}
\usepackage{multirow}
\usepackage{color,soul}
\usepackage{booktabs, makecell}

\begin{document}

\title{\LARGE \bf SimTO-Robust: Automated Design of Soft Grippers Robust to Grasp Pose and Object Geometry}

\author{
    Kurt Enkera$^{1,2}$, Joshua Pinskier$^{1}$, Marcus Gallagher$^{2}$, David Howard$^{1}$
    \thanks{
        $^{1}$ CSIRO Robotics, Data61, CSIRO, 
        Australia. 
        E-mails: {\tt\footnotesize \emph{firstname.lastname}@csiro.au}
    }
    \thanks{
        $^{2}$ School of Electrical Engineering and Computer Science (EECS), The University of Queensland, St Lucia, Australia.
    }
    \thanks{ $^{*}$ This work was supported by the CSIRO's Data61 AI4Design portfolio.}
}
 
\bstctlcite{IEEEexample:BSTcontrol}

\maketitle

\input{chapters/00-abstract}
\input{chapters/01-introduction}
\input{chapters/02-method}
\input{chapters/03-numerical_results}

\input{chapters/04-real_validation}
\input{chapters/05-conclusion}

\balance{}
\bibliographystyle{./bibliography/IEEEtran}
\bibliography{ref}
\end{document}

%% file: chapters/00-abstract.tex
\begin{abstract}
Robotic grippers operating in unstructured environments must robustly grasp objects of different shapes and sizes despite perception and positioning errors. Soft grippers that can achieve this robustness through intelligent mechanical design rather than sophisticated control software offer a promising solution. However, no existing computational design framework explicitly optimises soft gripper morphology for robustness across object geometries and grasp poses. We introduce \emph{SimTO-Robust}, a generalisation of simulation-driven topology optimisation (\emph{SimTO}) that simulates grasps across multiple objects and poses and aggregates their contact forces to guide gripper morphology optimisation. We generate 6,646 designs and physically evaluate 24 of them alongside 10 specialised SimTO designs, two rigid baseline grippers, and a baseline generalist soft gripper produced using conventional topology optimisation. All 37 grippers are tested in a pick-and-place task involving five feature-rich objects and six grasp poses, with external disturbances of increasing magnitude applied during the task. On average, SimTO-Robust designs complete pick-and-place more reliably than specialised SimTO designs and the rigid baselines. The best SimTO-Robust designs also withstood more of the applied disturbances than any other tested gripper. These results show that incorporating diverse grasp conditions into simulation-driven topology optimisation is a simple and effective approach to improving real-world soft gripper robustness. More broadly, they show that robust robotic grasping can be achieved through intelligent mechanical design rather than sophisticated sensing and control.
\end{abstract}

%% file: chapters/01-introduction.tex
\section{Introduction} \label{sec:intro}

While robotic grasping has become ubiquitous in logistics and is rapidly expanding into unstructured settings such as fruit harvesting \cite{pinskier2021_frombioinsp}, coral farming \cite{pinskier2026reefflex} and household cleaning \cite{black2024pi0}, its reliance on visually tracking defined object classes limits both its success and \emph{robustness}. In unstructured environments, objects vary widely in shape and size, making it difficult for perception-based grasping systems to recognise and accurately localise every object encountered \cite{LI2024102792}. Even when objects are successfully detected, positioning errors can introduce uncertainty in grasp pose. Therefore, to enable the automation of laborious manual handling tasks in unstructured environments, robotic grippers must be \emph{robust} to variations in object geometry and grasp pose.

Soft robotic grippers offer a promising approach to robust grasping. Unlike conventional rigid grippers, which have limited passive adaptability and often rely on sophisticated vision-based control algorithms to achieve robustness~\cite{LI2024102792, black2024pi0}, soft grippers made from compliant materials can deform around objects and passively adapt to different object geometries and grasp poses~\cite{pinskier2021_frombioinsp}, reducing the reliance on complex sensing and control through \emph{embodied intelligence}~\cite{hughes2022embodied}. However, designing soft grippers is challenging as their complex material properties and continuous deformability give rise to high-dimensional design spaces~\cite{pinskier2021_frombioinsp}.


\begin{figure}[t!]
    \centering
    \includegraphics[width=0.46\textwidth]{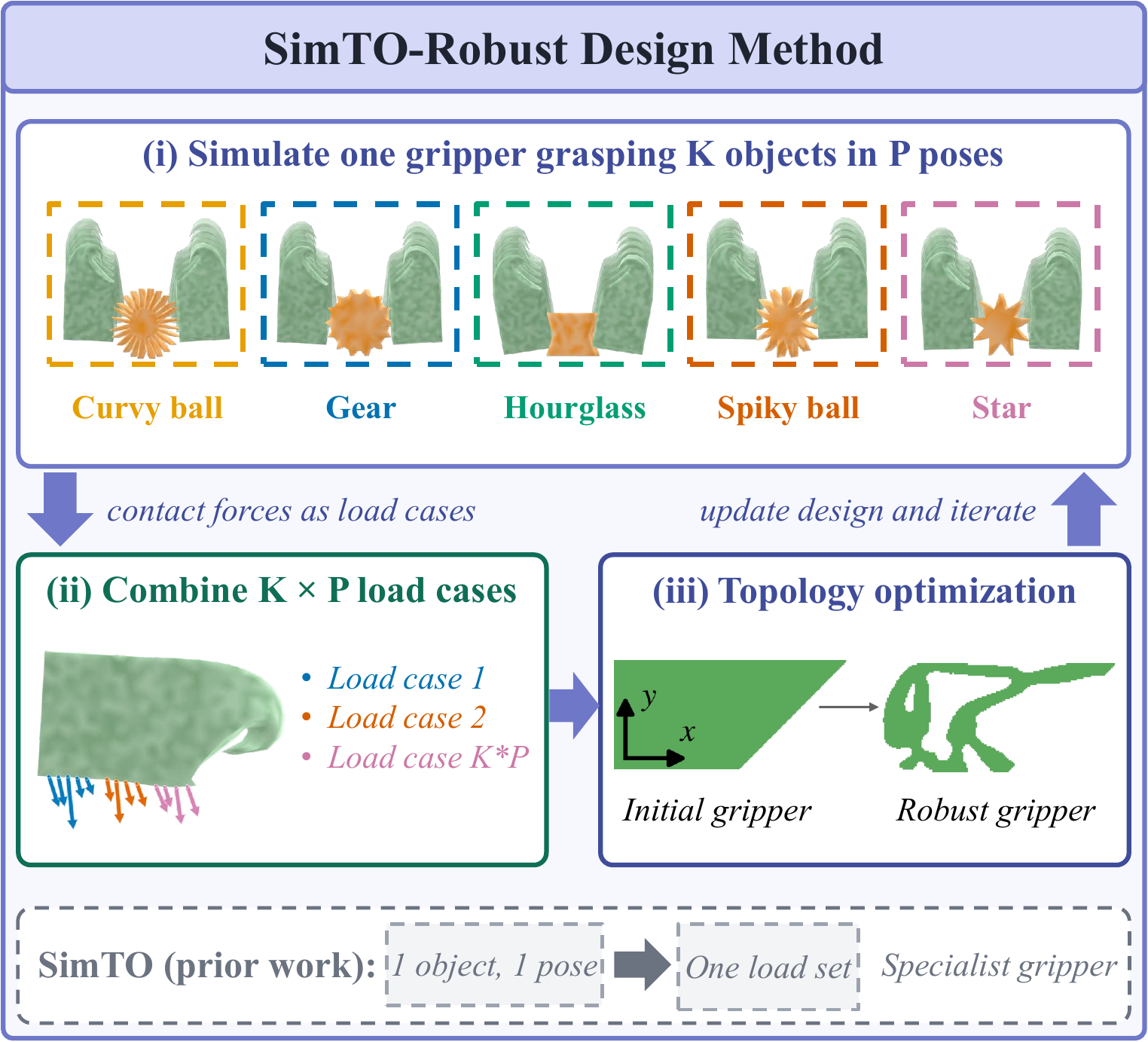}
    \caption{Whereas SimTO \cite{enkera2026simto} simulates a gripper grasping a single object in a single pose, SimTO-Robust (i) simulates grasps of $K$ objects in $P$ poses, (ii) combines the resulting contact forces, and (iii) uses them to update the gripper through topology optimisation. Iterating steps (i)-(iii) produces grippers with larger workspaces, grooved contact surfaces and other diverse features for more robust grasping than specialised SimTO designs and baseline rigid and soft grippers (Sec.~\ref{sec:realresults}).}
\label{fig:fig1}
\end{figure}


Soft gripper design methods range from manual, bioinspired approaches~\cite{Pfaff2011FinRay, pneunet2011} to highly automated computational design methods~\cite{pinskier2021_frombioinsp}. Manual approaches rely heavily on designer intuition and repeated prototyping, and become increasingly difficult as the number of target objects and grasp poses grows. Computational design methods, such as parametric and topology optimisation (TO), use physics-based simulations to systematically explore and evaluate a larger range of candidate designs than manual approaches~\cite{pinskier2021_frombioinsp}. However, parametric optimisation requires designers to specify an initial geometry (e.g., a Fin Ray finger) and an often limited set of parameters to optimise (e.g., finger length or width), heavily constraining the range of achievable morphologies~\cite{xie2024finqd}. In contrast, TO does not require a predefined parameterised geometry, instead optimising the distribution of material within a general design domain to achieve a specified objective, producing more freeform morphologies with fine-grained geometric features~\cite{liu2018_TObenchmark_v2, pinskier2024}.

Despite their ability to efficiently search large design spaces, to our knowledge, no existing computational method explicitly optimises soft grippers for robustness across a specific set of object geometries and grasp poses. Instead, parametric approaches typically optimise grasp strength~\cite{liu2022simulation, Navez2024, wang2024finbayes}, while TO frameworks optimise generic flexibility-based objectives~\cite{liu2018_TObenchmark_v2, wang2020topology, caasenbrood2020, liu2023, pinskier2024}. In both cases, robustness to variations in object geometry and grasp pose is not explicitly represented in the design objective.

TO is a promising candidate for robust soft gripper design as it produces freeform morphologies while requiring less human input than parametric approaches. However, expressing grasp robustness, which depends on contact-rich interactions across many objects and poses, as a tractable optimisation objective remains challenging. Simulation-driven topology optimisation, or \emph{SimTO} \cite{enkera2026simto}, offers a solution for a single object in a single pose. It simulates soft grasping and uses the resulting gripper-object contact forces to define an object-specific, flexibility-based objective. However, as it only models one object and pose, the resulting designs are \emph{specialised} to that object-pose combination and lack robustness.

To improve the robustness of SimTO, we develop \emph{SimTO-Robust}, which simulates soft grasping of \textbf{multiple objects and multiple poses} and aggregates the resulting contact forces into a single flexibility-based objective that accounts for every simulated grasp. The designer can therefore control what the gripper is robust to simply by choosing which objects and poses are simulated. Our contributions are:
\begin{itemize}
    \item A generalised TO framework which explicitly incorporates variations in object geometry and grasp pose into the topology optimisation of robust soft grippers;
    \item A dataset of 6,646 robust designs generated with SimTO-Robust, and an analysis of how their geometric features differ from those of specialised SimTO designs.
\end{itemize}

%% file: chapters/02-method.tex
\section{Automated Design Method}\label{sec:method}

SimTO-Robust generalises the SimTO computational design method \cite{enkera2026simto}. Whereas SimTO produces grippers specialised for a single object-pose combination, SimTO-Robust optimises grippers for robust grasping across multiple objects and poses. Designers can select these objects and poses to reflect their target application (e.g., representative fruit geometries and grasp poses for a robust fruit-picking gripper).

Before optimisation begins, the designer must specify (1) a design domain within which candidate gripper morphologies are generated, and (2) a fixed actuation mechanism that can be implemented both in simulation and reality. Starting from an initial gripper within this domain, SimTO-Robust iteratively (i) simulates the gripper grasping $K$ objects across $P$ poses, (ii) combines the resulting $K \times P$ sets of gripper-object contact forces into a single load set, and (iii) uses these loads to update the gripper geometry through TO (Fig.~\ref{fig:fig1}). The updated gripper is then simulated again, producing new contact forces that guide further changes to its morphology.

Gripper optimisation is performed in 2D while grasping simulations are conducted in 3D. We use the Solid Isotropic Material with Penalisation (SIMP) topology optimisation method \cite{BendsoeSigmund2004}, which assigns a continuous material density $\rho_e \in [0,1]$ to each element of a 2D finite element method (FEM) mesh of the design domain, where $\rho_e = 1$ represents solid material and $\rho_e = 0$ represents void. To construct a 3D tetrahedral mesh for dynamic grasping simulations, the density field is first binarised by setting elements with $\rho_e \geq 0.5$ to solid and all others to void. The resulting binary design is then extruded and tetrahedralised for 3D simulation.

\subsection{SimTO-Robust Stages}

\emph{(i) Dynamic grasping simulations.} In this stage, the current gripper is simulated grasping $K$ objects in $P$ grasp poses using Taccel~\cite{taccel2025}, a GPU-parallelised soft robotics simulator. Each simulation begins with the gripper positioned in one of the $P$ poses relative to one of the $K$ objects. The gripper is then actuated to deform around the object, and then lifted.

The goal of each simulation is to extract gripper-object contact forces for 2D topology optimisation in stage (iii). At each simulation timestep, contact forces $\bm{f} \in \mathbb{R}^3$ acting on the surface of the deformed gripper mesh are mapped into the undeformed frame shown in Fig.~\ref{fig:fig1}(iii) using the Kabsch algorithm~\cite{kabsch1976}. Only forces with (1) a negative $y$-component and (2) a positive $x$-component in the undeformed frame are retained. These criteria isolate forces that apply a positive normal force to the object while resisting slip (i.e., forces that contribute to a secure grasp). If multiple forces are recorded at the same contact point during a simulation, only the force with the largest $x$-component is retained.

To project the retained 3D forces into 2D, the contact point coordinates are first rounded to align with the nodes of the 2D finite element mesh used in TO. Forces sharing the same $(x,y)$ coordinates are then averaged, and the $z$-component of each averaged vector is set to zero, yielding a 2D force $\bm{f} \in \mathbb{R}^2$ at the corresponding $(x,y)$ location. Each simulation therefore produces a filtered set of 2D contact forces, yielding $K \times P$ load sets in total.

\emph{(ii) Load combination.} To include contact forces from all $K \times P$ simulations in topology optimisation, we combine the load sets from stage (i) into a single load set. Although the forces within each load set act at distinct nodes, forces from different simulations may act at the same node. In such cases, we retain only the force with the largest $x$-component, mirroring the rule used for repeated contacts within each simulation in stage (i). If only one set contains a force at a given node, that force is retained. The result is a single load set $\{\bm{f}_1, \ldots, \bm{f}_{N_f}\}$ with at most one force per node, which is passed to TO in stage (iii).


\begin{figure*}[t!]
    \centering
    \includegraphics[width=0.95\textwidth]{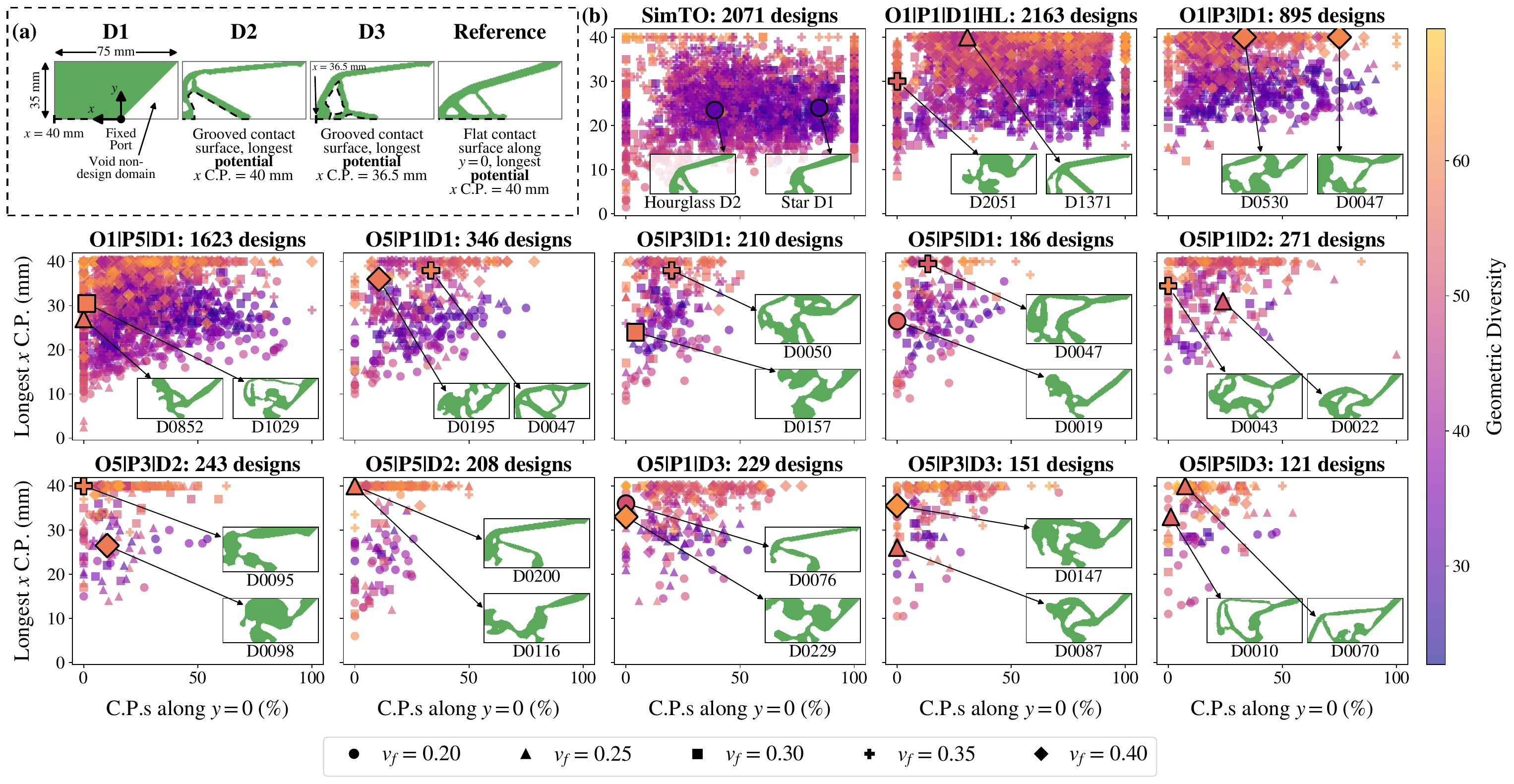}
    \caption{\emph{Numerical optimisation results.} \textbf{a)} The three initial designs used in the SimTO-Robust experiments, and the reference generalist design from \cite{liu2018_TObenchmark_v2}. \textbf{(b)} The 12 SimTO-Robust experiments plus the SimTO baseline (\ex{O1|P1|D1}). Each marker is one generated design, and we plot its furthest simulated contact point (C.P.) in $x$ from the fixed port against the percentage of its C.P.s lying on $y=0$. Outlined, enlarged markers are designs that were 3D-printed and validated in this work.}
    \label{fig:optimization}
\end{figure*}

\emph{(iii) Topology optimisation.} The combined load set from (ii) is used to update the density vector $\bm{\rho}$ using the flexibility-based objective of SimTO~\cite{enkera2026simto}. This objective encourages the gripper to deform in the direction of the simulated contact forces \textbf{$\{\bm{f}_1, \ldots, \bm{f}_{N_f}\}$}. By incorporating contact forces across multiple objects and poses, the optimisation favours morphologies that accommodate a broader range of grasp conditions than single-object, single-pose SimTO. The updated design is then binarised and tetrahedralised, and simulated in stage (i) of the next iteration. Optimisation terminates after 20 iterations or when two successive designs, $\bm{\rho}^i$ and $\bm{\rho}^{i+1}$, satisfy $\sum_e(\rho_e^{i+1} - \rho_e^{i})^4 < 10$.

%% file: chapters/03-numerical_results.tex
\section{Numerical Experiments}\label{sec:simresults}

We performed twelve SimTO-Robust optimisation experiments, generating 6,646 robust designs in total (Fig.~\ref{fig:optimization}(b)). Each experiment is labelled using a code of the form \ex{O$K$|P$P$|D$n$}, where $K$ is the number of simulated objects, $P$ is the number of simulated grasp poses per object and $n \in \{1,2,3\}$ enumerates the three different initial designs in Fig.~\ref{fig:optimization}(a). Our design domain is defined as the green trapezoidal region shown in \ex{D1} (Fig.~\ref{fig:optimization}(a)). 

\ex{O1} experiments simulated a single object and were repeated for each of the five feature-rich objects in Fig.~\ref{fig:fig1}, whereas \ex{O5} experiments simulated all five objects in each run. The grasp poses used in \ex{P1}, \ex{P3} and \ex{P5} experiments differed only in $z$-height. This is because pose offsets in $x$, $y$, or $\theta_z$ (rotation about the $z$-axis) can produce out-of-plane contact forces that contribute to grasp stability during simulation. However, our 2D TO formulation discards the out-of-plane components of all contact forces, so we expect these offsets to have a limited effect on optimised morphologies, hence we did not include them in our experiments.

We also performed a special \ex{HL} (\emph{high-low}) experiment, denoted \ex{O1|P1|D1|HL}, in which the gripper was positioned higher in $z$ during the first iteration and lower in subsequent iterations, whereas standard \ex{P1} experiments maintained a fixed grasp height across iterations. The three initial designs shown in Fig.~\ref{fig:optimization}(a) include \ex{D1}, which completely fills the design domain with material, and two optimised designs, \ex{D2} and \ex{D3}, generated by the \ex{HL} experiment.

To generate a diverse set of designs, we swept two parameters: the target volume fraction $v_f$, which specifies the proportion of the design domain occupied by solid material, and the Young's modulus of simulated objects $E_o$. All \ex{O5} experiments swept five volume fractions, $v_f \in \{0.20, 0.25, 0.30, 0.35, 0.40\}$, and twelve object moduli, $E_o \in \{1,2,\ldots,12\}\,\text{MPa}$, with the gripper modulus fixed at $E_g = 12\,\text{MPa}$, yielding 60 optimisation runs per experiment. For \ex{O1} experiments, the same sweep was repeated for each of the five objects, yielding 300 optimisation runs per experiment and consequently more generated designs.

\subsection{Analyzing the Contact Surfaces of Robust Designs}\label{sec:contactsurfaces}

In Fig.~\ref{fig:optimization}(b), we plot two measures of every design's contact surface. Both measures are computed from the combined load set $\{\bm{f}_1,\ldots,\bm{f}_{N_f}\}$ obtained by simulating the design (i.e., the loads that would drive its next round of TO). Each load in this set acts at a contact point (C.P.) with coordinates $(x, y)$ in the 2D design domain, expressed in the coordinate frame drawn on \ex{D1} in Fig.~\ref{fig:optimization}(a). The first measure is the largest $x$-coordinate among a design's contact points, which indicates how far from the fixed port its contact with the object(s) extends. This value is bounded by the length of the design's contact surface, marked in Fig.~\ref{fig:optimization}(a) as the longest potential $x$ C.P. (40~mm for \ex{D1} and \ex{D2}, 36.5~mm for \ex{D3}). The second measure is the percentage of a design's contact points that lie on the line $y = 0$, which is high for a flat contact surface and low for a grooved one.

For the 2,071 SimTO designs, the mean longest $x$ C.P. was 24.8~mm, compared with 32.4~mm for the 6,646 SimTO-Robust designs. This indicates that, on average, SimTO produced designs with shorter contact surfaces, such as the two SimTO designs displayed in Fig.~\ref{fig:optimization}(b). Further to this, 56.81\% of the SimTO contacts were located along $y = 0$, compared with only 33.70\% for SimTO-Robust. This indicates that the robust designs generally had more grooved contact surfaces than the specialised designs. Consistent with this, the average $y$ C.P. (the mean height of the contact points above $y=0$) was 5.4~mm for SimTO-Robust designs and 3.4~mm for SimTO, showing that contact occurred deeper within the grooves for robust designs. These grooves are visible in the 24 SimTO-Robust designs shown in Fig.~\ref{fig:optimization}(b). 

Each design in Fig.~\ref{fig:optimization}(b) is also coloured by its geometric diversity, computed for the $i$th design as $D_i = \| \bm{\rho}_i - \bar{\bm{\rho}}\|$, where $\bm{\rho}_i$ is the $150\times70=10{,}500$-dimensional binary vector representing design $i$. The mean morphology $\bar{\bm{\rho}}$ is obtained by averaging the 8,717 designs produced across the SimTO and SimTO-Robust experiments, giving a density field in $[0,1]$, which is then binarized as in Section~\ref{sec:method}. The SimTO-Robust designs had an average geometric diversity of 46.0, compared with 40.1 for SimTO, indicating that the longer, more grooved contact surfaces of the robust designs were realised through a wider variety of morphologies.

\begin{table}[!tb]
\centering
\fontsize{8.5pt}{10.5pt}\selectfont
\setlength{\tabcolsep}{3.5pt}
\caption{Simulated grasp robustness success (\%) per test pose, averaged over all five objects and all five impulse directions. $N_{\text{d}}$: number of designs. Mean: average robustness over all six poses. The best value per column is emboldened.}
\label{tab:pose_table}
\begin{tabular}{lrccccccc}
\toprule
Design group & $N_{\text{d}}$ & $O$ & $Z_1$ & $Z_2$ & $X_1$ & $Y_1$ & $\theta_1$ & Mean \\
\midrule
\ex{O1|P1|D1} & 2071 & 92.3 & 77.2 & 43.3 & 76.7 & 36.6 & 75.8 & 67.0 \\
\ex{O1|P3|D1} & 895 & 93.0 & 89.0 & 68.5 & 89.1 & 34.0 & 86.5 & 76.7 \\
\ex{O1|P5|D1} & 1623 & 93.5 & 84.2 & 61.1 & 83.7 & 41.2 & 82.8 & 74.4 \\
\ex{O1|P1|D1|HL} & 2163 & \textbf{94.0} & \textbf{89.5} & \textbf{73.6} & \textbf{89.7} & 39.8 & \textbf{87.7} & \textbf{79.0} \\
\midrule
\ex{O5|P1|D1} & 346 & 92.7 & 83.1 & 61.7 & 83.3 & 44.8 & 82.7 & 74.7 \\
\ex{O5|P3|D1} & 210 & 90.5 & 80.9 & 60.5 & 80.3 & \textbf{45.8} & 80.0 & 73.0 \\
\ex{O5|P5|D1} & 186 & 87.4 & 78.8 & 56.5 & 78.4 & 44.2 & 77.5 & 70.5 \\
\bottomrule
\end{tabular}
\end{table}

\subsection{Effect of the Number of Objects, Poses and Initial Design}\label{sec:abl_KP}

We evaluated all designs from Fig.~\ref{fig:optimization}(b) in simulation on all five feature-rich objects in six test poses: a nominal pose ($O$); the nominal pose raised by $7\,\text{mm}$ and $14\,\text{mm}$ in $z$ ($Z_1$, $Z_2$); and the $Z_1$ pose offset by $8\,\text{mm}$ in $x$ ($X_1$), offset by $4\,\text{mm}$ in $y$ ($Y_1$), and rotated by $8^{\circ}$ about $z$ ($\theta_1$). For each object-pose combination, the gripper grasps and lifts the object, after which a $60\,\text{N}$ force is applied briefly in each of the five directions in Fig.~\ref{fig:exp_setup}(d), giving $5 \times 6 \times 5 = 150$ grasps per design. A grasp is successful if the object is lifted and still held after the disturbance, and we refer to the percentage of successful grasps as the \emph{grasp robustness success}.

\subsubsection{Effects of Increasing Pose Count}\label{sec:posecount}

\begin{figure}[t!]
    \centering
    \includegraphics[width=0.38\textwidth]{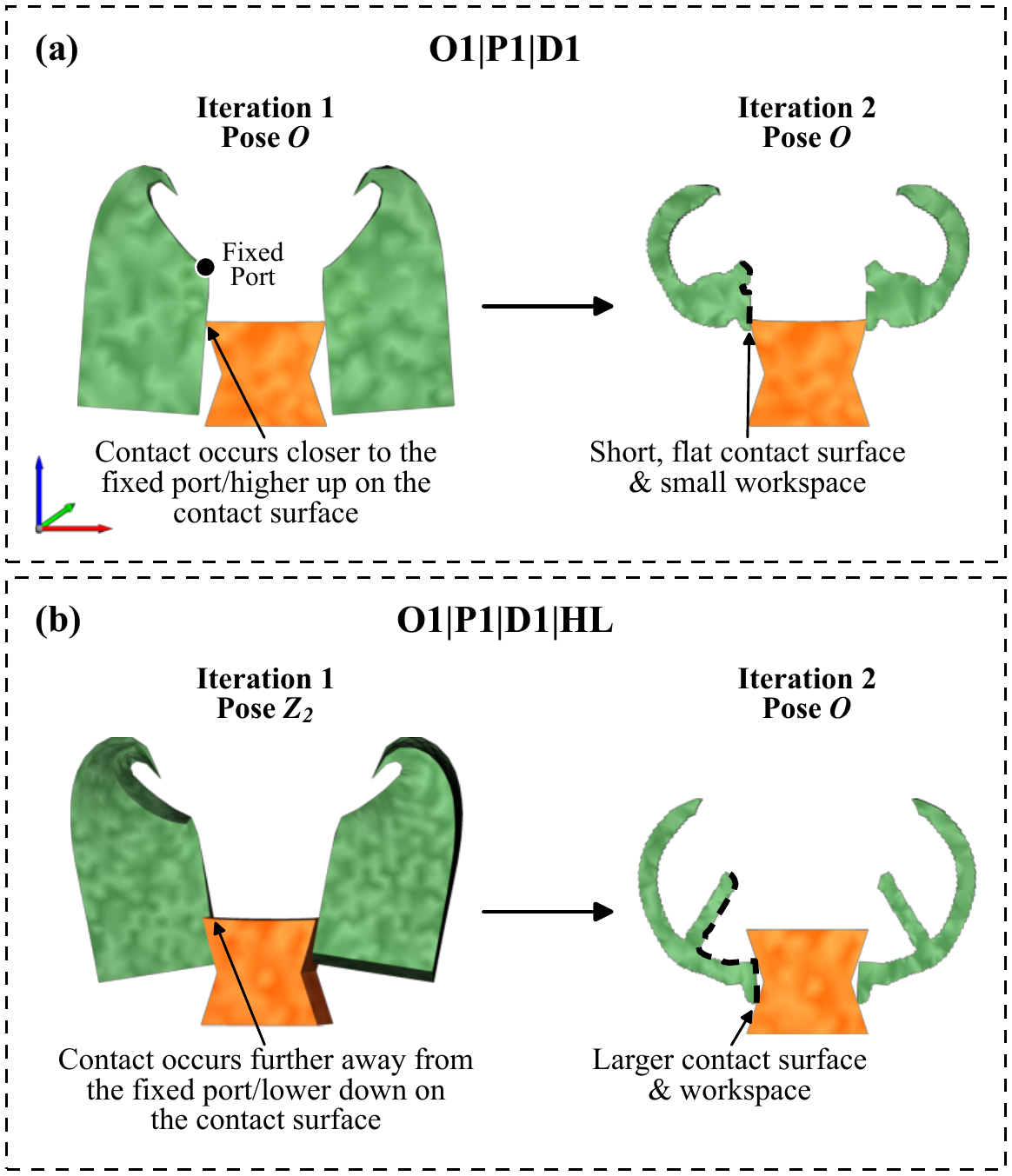}
    \caption{\emph{Where contact occurs on the gripper shapes the optimised design.} Simulated deformation of the initial and subsequent designs during optimisation of an hourglass specialist in \textbf{(a)} \ex{O1|P1|D1} and \textbf{(b)} \ex{O1|P1|D1|HL}.} 
    \label{fig:HL_comparison}
\end{figure}

Table~\ref{tab:pose_table} shows that SimTO designs (\ex{O1|P1|D1}) are highly sensitive to grasp height. They succeed in 92.3\% of grasps at $O$, the pose they were optimised at, but only 77.2\% at $Z_1$ and 43.3\% at $Z_2$. These results arise because our optimiser showed a strong tendency to place material where contact occurs and remove material elsewhere. During optimisation, all \ex{O1|P1|D1} grasps were simulated at the lowered nominal pose $O$, causing contact to mostly occur very close to the fixed port (Fig.~\ref{fig:HL_comparison}(a)). Material further along the contact surface was therefore often removed and rarely recovered in subsequent iterations. On average, this produced designs with shorter contact surfaces (Section~\ref{sec:contactsurfaces}) that grasp well at $O$ but fail at higher poses. Optimising over three poses (\ex{O1|P3|D1}) supplies contact forces along more of the surface, producing designs with longer, grooved contact surfaces and raising success at $Z_2$ from 43.3\% to 68.5\% and overall success from 67.0\% to 76.7\%. However, increasing to five poses gives no further gain.

\ex{O1|P1|D1|HL} is the most robust group in Table~\ref{tab:pose_table}, achieving the best overall success (79.0\%) despite using only one pose per iteration. During optimisation, \ex{O1|P1|D1|HL} designs were simulated in the raised pose $Z_2$ in the first iteration and the lowered nominal pose $O$ thereafter. Because grasping from $Z_2$ produces contact far from the fixed port, the first optimised design tended to have a long contact surface (Fig.~\ref{fig:HL_comparison}(b)). In subsequent iterations, this surface continued to contact the object even in the lowered pose, with object features often making contact deep within its grooves. The optimiser therefore tended to preserve the long contact surface while refining its shape. As a result, \ex{O1|P1|D1|HL} outperforms \ex{O1|P3|D1} at one third of its per-iteration simulation cost. This shows that varying the simulated pose across iterations, rather than simulating many poses per iteration, offers a cheaper route to robust designs.

\begin{figure}[t!]
    \centering
    \includegraphics[width=0.4\textwidth]{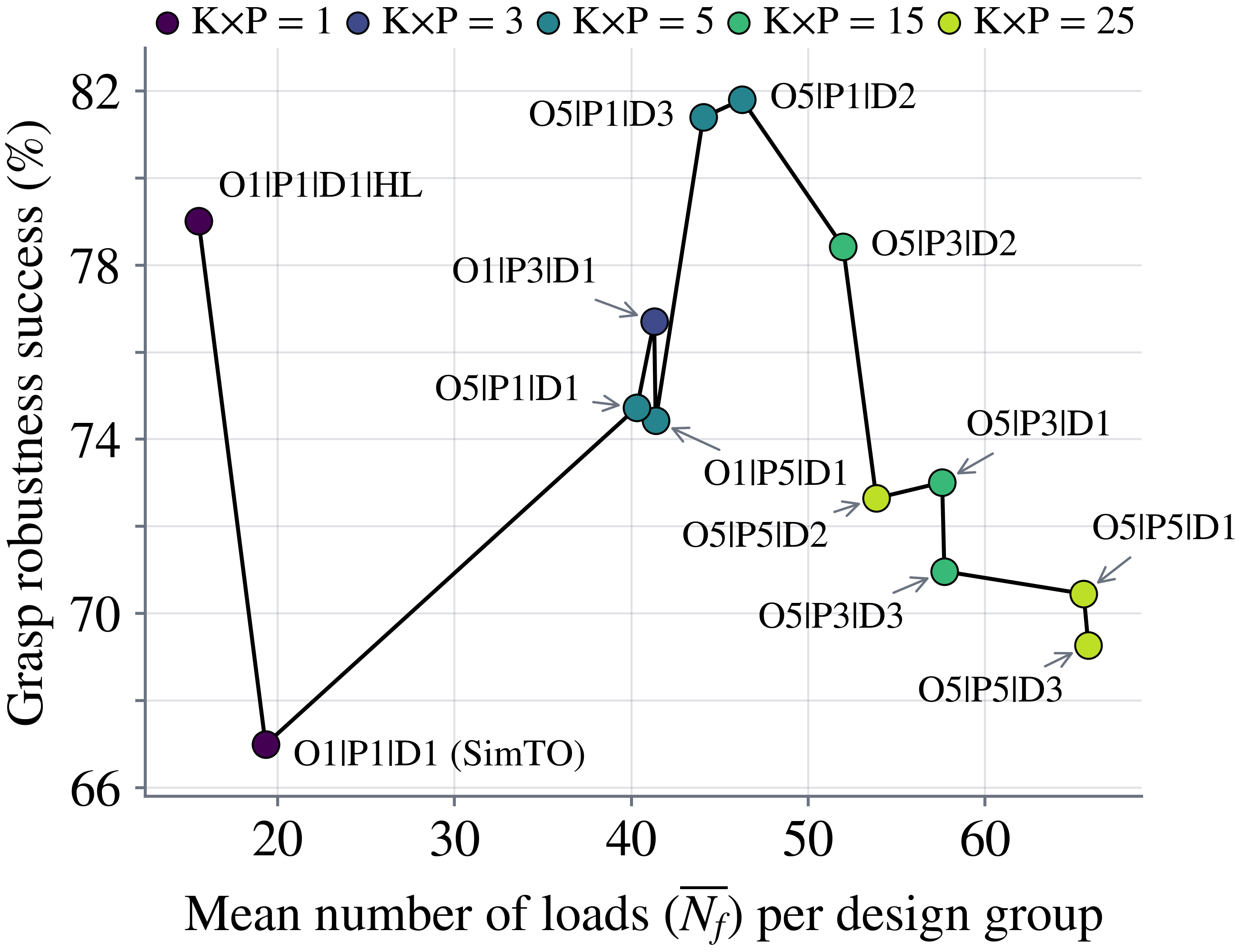}
    \caption{\emph{Trade-off between grasp robustness and optimisation cost.} Simulating more objects and poses increases the mean load-set size $\overline{N_f}$ and the computational cost. Grasp robustness, our measure of embodied intelligence, peaks at approximately 45 loads ($K \times P = 5$) and degrades beyond this point.}
    \label{fig:knee}
\end{figure}

For the multi-object generalists (\ex{O5}), adding poses does not improve robustness, with success declining slightly from 74.7\% (\ex{O5|P1|D1}) to 73.0\% (\ex{O5|P3|D1}) and 70.5\% (\ex{O5|P5|D1}). The same trend holds for the other initial designs, with \ex{O5|P1|D2} and \ex{O5|P1|D3} declining from 81.8\% and 81.4\% to 72.6\% and 69.3\% at five poses. As Fig.~\ref{fig:knee} shows, each added pose substantially increases the mean load-set size $\overline{N_f}$ used in topology optimisation. This suggests that for multi-object generalists, whose five objects already supply rich contact information, additional poses can dilute this information and yield less robust designs.

\begin{table}[!t]
\centering
\fontsize{8.5pt}{10.5pt}\selectfont
\setlength{\tabcolsep}{3.5pt}
\caption{Simulated grasp robustness success (\%) per test object, averaged over all poses and impulse directions, for the \ex{D1} groups in Table~\ref{tab:pose_table}. \emph{Matched}: specialists ($K=1$) optimised for that object. \emph{Other}: the four remaining object specialists in the group. \emph{Gen.}: the generalist ($K = 5$) with the same $P$. Mean: average robustness over all objects. For each $P$, the best value per column is emboldened.}
\label{tab:object_table}
\begin{tabular}{llcccccc}
\toprule
$P$ & Designs & Curvy & Gear & Hourgl. & Spiky & Star & Mean \\
\midrule
\multirow{3}{*}{P1} & Matched & 60.4 & 79.0 & 39.1 & 57.9 & \textbf{83.3} & 63.9 \\
 & Other & 64.5 & 75.0 & 60.9 & 56.5 & 77.7 & 66.9 \\
 & Gen. & \textbf{73.4} & \textbf{80.6} & \textbf{66.8} & \textbf{70.4} & 82.5 & \textbf{74.7} \\
\midrule
\multirow{3}{*}{P3} & Matched & 72.5 & \textbf{84.2} & 66.2 & 66.8 & \textbf{88.6} & 75.7 \\
 & Other & \textbf{72.7} & 80.8 & \textbf{76.4} & \textbf{69.8} & 85.6 & \textbf{77.0} \\
 & Gen. & 72.6 & 77.7 & 65.2 & 68.7 & 80.8 & 73.0 \\
\midrule
\multirow{3}{*}{P5} & Matched & \textbf{72.1} & \textbf{84.7} & 61.0 & 64.4 & \textbf{87.0} & 73.9 \\
 & Other & 71.6 & 78.6 & \textbf{70.3} & \textbf{69.2} & 83.2 & \textbf{74.6} \\
 & Gen. & 69.5 & 73.9 & 63.7 & 65.9 & 79.2 & 70.5 \\
\midrule
\multirow{2}{*}{P1$|$HL} & Matched & 73.5 & \textbf{86.1} & 73.7 & \textbf{76.2} & 83.7 & 78.6 \\
 & Other & \textbf{73.7} & 83.7 & \textbf{83.9} & 69.6 & \textbf{85.9} & \textbf{79.4} \\
\bottomrule
\end{tabular}
\end{table}

\subsubsection{Effect of Object Count}\label{sec:objectcount}

To investigate the effect of object count, we ask three questions: how well do specialists ($K=1$) grasp the object they were optimised for, how well do they grasp other objects, and how do multi-object generalists ($K=5$) compare with specialists? For a fair comparison, we only analyse the seven \ex{D1} groups in Table~\ref{tab:pose_table}. For each test object, Table~\ref{tab:object_table} compares the specialists optimised for that object (\emph{Matched}), the specialists optimised for the other four objects (\emph{Other}), and the generalist with the same $P$ (\emph{Gen.}).

Table~\ref{tab:object_table} shows that specialists can achieve high grasp robustness not only on the objects they were optimised for, but also on other objects. Across the 20 object-pose combinations in Table~\ref{tab:object_table}, matched specialists achieve the highest success in eight, and specialists optimised for other objects also achieve the highest success in another eight. This suggests that, for our feature-rich object set, optimising for a single object can produce morphological features that generalise to other objects. The generalists achieve the highest success in the remaining four combinations, all at \ex{P1}, where they outperform both matched and other specialists on every object except the star. At \ex{P3} and \ex{P5}, however, specialists slightly outperform generalists, indicating that multi-object optimisation helps most when only one pose is simulated.

\subsubsection{Robustness versus Cost}\label{sec:robustnesscost}

Fig.~\ref{fig:knee} plots each design group's grasp robustness success against its mean load-set size $\overline{N_f}$. Robustness increases with $\overline{N_f}$ up to an apparent sweet spot at roughly 45 loads, beyond which it declines. Below this point, too few objects or poses are simulated to capture the contact variety needed for robust designs. However, above this point, additional loads appear to dilute rather than enrich the contact information. \ex{O1|P1|D1|HL} is a notable exception. By varying just a single simulated pose across iterations, it achieves high robustness with the smallest average load set size of any group.

The computational cost of each optimisation iteration scales with the number of simulations, $K \times P$. In our experiments, a single Taccel simulation took up to 30 minutes, so with at most 20 iterations, a single optimisation run requires up to $10 \times K \times P$ simulation-hours. The sweet spot coincides with $K \times P = 5$, beyond which both cost and robustness worsen. Although these findings are specific to our object set and 2D formulation, they show that increasing the number of simulated objects and poses beyond a certain point can increase computational cost without improving robustness.

\subsubsection{Effect of Initial Design} Fig.~\ref{fig:knee} also shows that the initial design has a large effect. Multi-object generalists optimised with \ex{D2} and \ex{D3} as their initial design reach a grasp robustness success of 81.8\% and 81.4\% at \ex{P1}, compared with 74.7\% for \ex{D1}, and are the most robust groups in this study. Starting from an already robust design can therefore increase the benefit of multi-object optimisation. This advantage is reduced, however, as more poses are added.

%% file: chapters/04-real_validation.tex
\section{Experimental Validation}\label{sec:realresults}

\begin{figure*}[t!]
    \centering
    \includegraphics[width=0.75\textwidth]{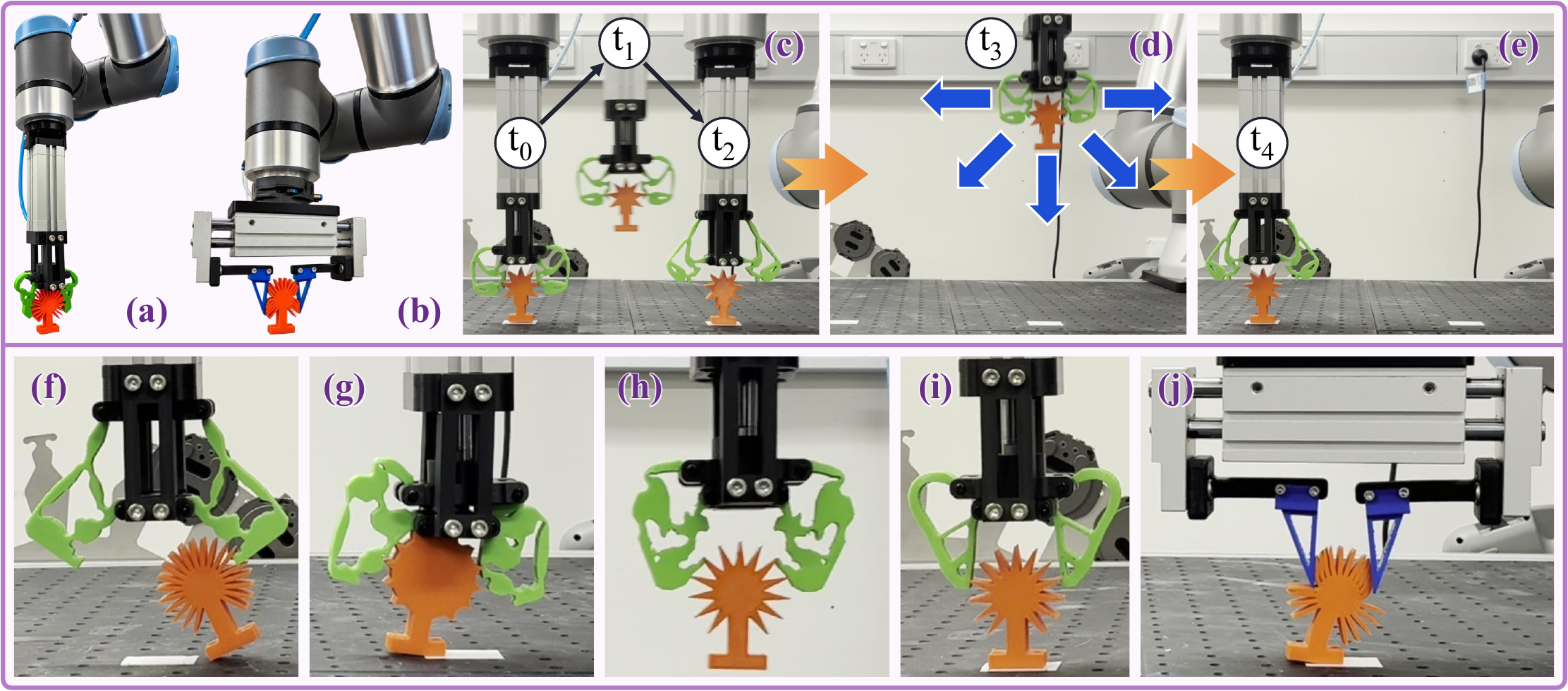}
    \caption{\emph{Experimental grasp robustness tests.} \textbf{(a):} Soft grippers were mounted on a UR10e arm and actuated by a linear pneumatic actuator. \textbf{(b):} Our parallel-jaw rigid gripper. \textbf{(c)-(e):} In each trial, the object is first grasped at $t_0$, transported during $t_1$ and placed at $t_2$. It is then re-grasped and lifted during $t_3$ before undergoing four impulse-resistance tests, where the impulses grow in magnitude. First, the gripper is accelerated in all five directions shown in blue. After, the gripper is shaken at three increasing accelerations. If the object is still held, it is placed back on the target at $t_4$. \textbf{Bottom row:} Representative failure modes: \textbf{(f)} the object sticking to gripper features; \textbf{(g)} asymmetric deformation at low $z$ grasp poses; \textbf{(h)} a robust design with a loose grip, causing failure during shaking; \textbf{(i)} the flat contact surface of the soft reference design fails to strongly grip objects; \textbf{(j)} a rigid gripper failing to place the object accurately in the $X_1$ pose.}
    \label{fig:exp_setup}
\end{figure*}

\begin{figure*}[ht!]
    \centering
    \includegraphics[width=\textwidth]{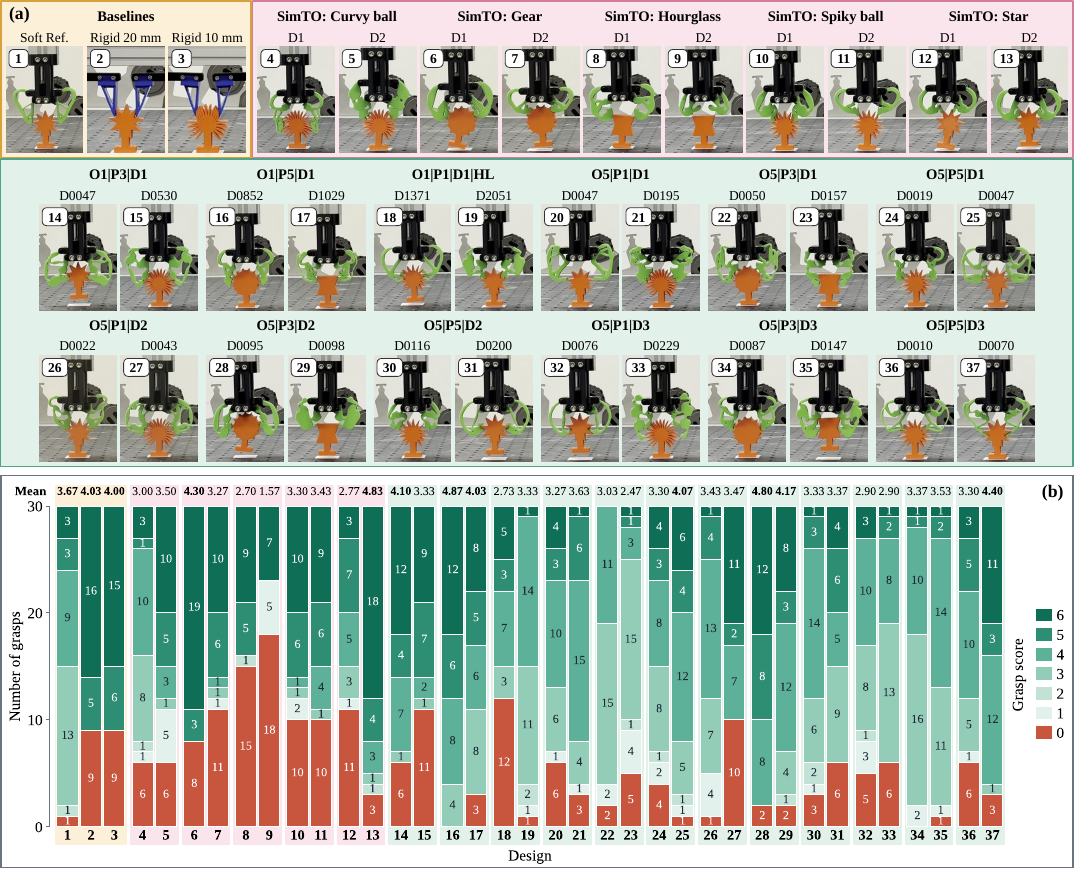}
    \caption{\emph{Experimental grasping results.} \textbf{(a)} All 37 validated grippers in their grasped state at pose $Z_2$, grouped into baselines (yellow), SimTO specialists (pink), and SimTO-Robust designs (green). \textbf{(b)} Grasp score distribution of each design over its 30 grasps. Mean scores equal to or greater than that of the soft reference design are in bold.}
    \label{fig:results}
\end{figure*}

We 3D-printed and tested the robustness of 37 grippers: two designs from each of the 12 SimTO-Robust experiments (the 24 designs in Fig.~\ref{fig:optimization}(b)), 10 SimTO designs, two rigid baseline grippers, and a soft reference design~\cite{liu2018_TObenchmark_v2}. The SimTO-Robust designs were hand-selected to capture a diverse range of morphologies, with priority given to designs in the upper-left region of the plots in Fig.~\ref{fig:optimization}(b) (i.e., those with long, grooved contact surfaces). The two rigid grippers (Fig.~\ref{fig:results}(a)) are standard parallel-jaw end-effectors 3D-printed in Acrylonitrile Styrene Acrylate (ASA), a stiff thermoplastic with a Young's modulus of approximately $2.4\,\text{GPa}$~\cite{bambu_asa_tds}. One is $10\,\text{mm}$ wide, matching the width of all soft fingers, and the other is $20\,\text{mm}$ wide. Both have a flat, $40\,\text{mm}$-long contact surface, which is equal to the maximum possible contact surface length of any soft design.

All grippers were mounted on a UR10e robot arm and evaluated on a physical pick-and-place task, during which external disturbances of increasing magnitude were applied (Fig.~\ref{fig:exp_setup}(c)-(e)). Each design performed the task with the five feature-rich objects in the six grasp poses from Section~\ref{sec:abl_KP}, giving 30 grasps per design. All soft grippers and objects were 3D-printed in NinjaFlex TPU~\cite{ninjatek_ninjaflex_tds}, with an experimentally measured Young's modulus of $12\,\text{MPa}$. Each object was printed with a $27\,\text{mm} \times 15\,\text{mm}$ rectangular base so that it could stand upright. At the start of each trial, the object was placed by hand at the center of a $37\,\text{mm} \times 25\,\text{mm}$ strip of white masking tape, and the gripper was tasked with placing it on an identical target strip $225\,\text{mm}$ away.

Each trial began with the arm at its assigned grasp pose and proceeded through six stages. First, \textbf{(1)} the gripper grasped the object, transported it, and placed it on the second white target strip. \textbf{(2)} The object was then grasped and lifted again, and accelerated at $12{,}000\,\text{mm/s}^2$ over $100\,\text{mm}$ in each of the five directions shown in Fig.~\ref{fig:exp_setup}(d). Next, the arm traced a square with $10\,\text{mm}$ sides in the horizontal plane five times (as in~\cite{pinskier2026reefflex}) at three increasing accelerations, $4{,}000$, $8{,}000$ and $12{,}000\,\text{mm/s}^2$, which we refer to as \textbf{(3)} low-, \textbf{(4)} medium- and \textbf{(5)} high-speed shaking. Finally, \textbf{(6)} if the object was still held, the gripper attempted to place it upright back on the starting strip. Each grasp was scored from 0 to 6 by the number of consecutive stages completed, so a score of 0 denotes a failed pick-and-place. The object was then returned by hand to its starting position for the next grasp. 

\subsection{Experimental Results}

Fig.~\ref{fig:results} shows the grasp score distribution of every design. On average, SimTO-Robust designs were considerably more reliable than specialised SimTO designs, completing the first round of pick-and-place in 86.2\% of grasps compared with 67.3\% for SimTO. Robust designs also achieved a higher mean grasp score of 3.55 (averaged over all 24 designs) compared with 3.27 for SimTO (averaged over 10 designs). Both rigid grippers achieved high mean grasp scores (4.00 and 4.03), as they rarely dropped an object once grasped, but each failed 30\% of pick-and-place attempts. Conversely, the soft reference failed only once, but its weak grip (Fig.~\ref{fig:exp_setup}(i)) resulted in a lower mean grasp score of 3.67. Seven SimTO-Robust designs and two SimTO designs matched or exceeded the best rigid gripper's mean grasp score. The best of these, design 16, never failed pick-and-place and achieved the highest mean grasp score of any design (4.87), exceeding both rigid grippers and the soft reference.

To compare how well the designs withstood external disturbances, we scored each grasp by the fraction of the four disturbances it withstood, counting a failed pick-and-place as zero. The 24 SimTO-Robust designs achieved an average disturbance rejection success rate of 63.0\%, compared with 56.7\% for SimTO. Designs 16 and 28 shared the best individual disturbance rejection rate of 86.7\%, while the best SimTO design (Star D2) achieved 83.3\%, both rigid grippers 70.0\%, and the soft reference design 65.0\%.

The SimTO designs and rigid grippers performed in an ``all-or-nothing'' manner. Figure~\ref{fig:results} shows that they either failed pick-and-place frequently or, once they had grasped the object, they held it through most of the disturbances. For example, 80\% of SimTO grasps scored either 0 or at least 5, and \textbf{every rigid grasp} scored either 0 (30\%) or at least 5 (70\%). The SimTO-Robust designs were more consistent, with 54\% of their grasps scoring 3 or 4, compared with 14\% for SimTO and none for the rigid grippers. SimTO-Robust designs therefore lose some performance under the most extreme disturbances but experience far fewer failures (13.8\% of grasps failed, compared with roughly 30\% for SimTO and the rigid grippers). Because these extreme disturbances far exceeded those of typical pick-and-place tasks, this consistency makes the SimTO-Robust designs better suited for general-purpose use than designs that excel in some grasps but fail outright in others.

%% file: chapters/05-conclusion.tex
\section{Discussion}\label{sec:discussion}

Although SimTO-Robust designs outperformed specialised SimTO designs on average, their physical performance varied considerably (Fig.~\ref{fig:results}). This suggests that our design optimisation method improves grasp robustness overall, but does not guarantee that every resulting morphology will perform well. Here, we investigate the geometric features that distinguish high-performing designs from less robust ones.

\begin{figure}[t!]
    \centering
    \includegraphics[width=0.4\textwidth]{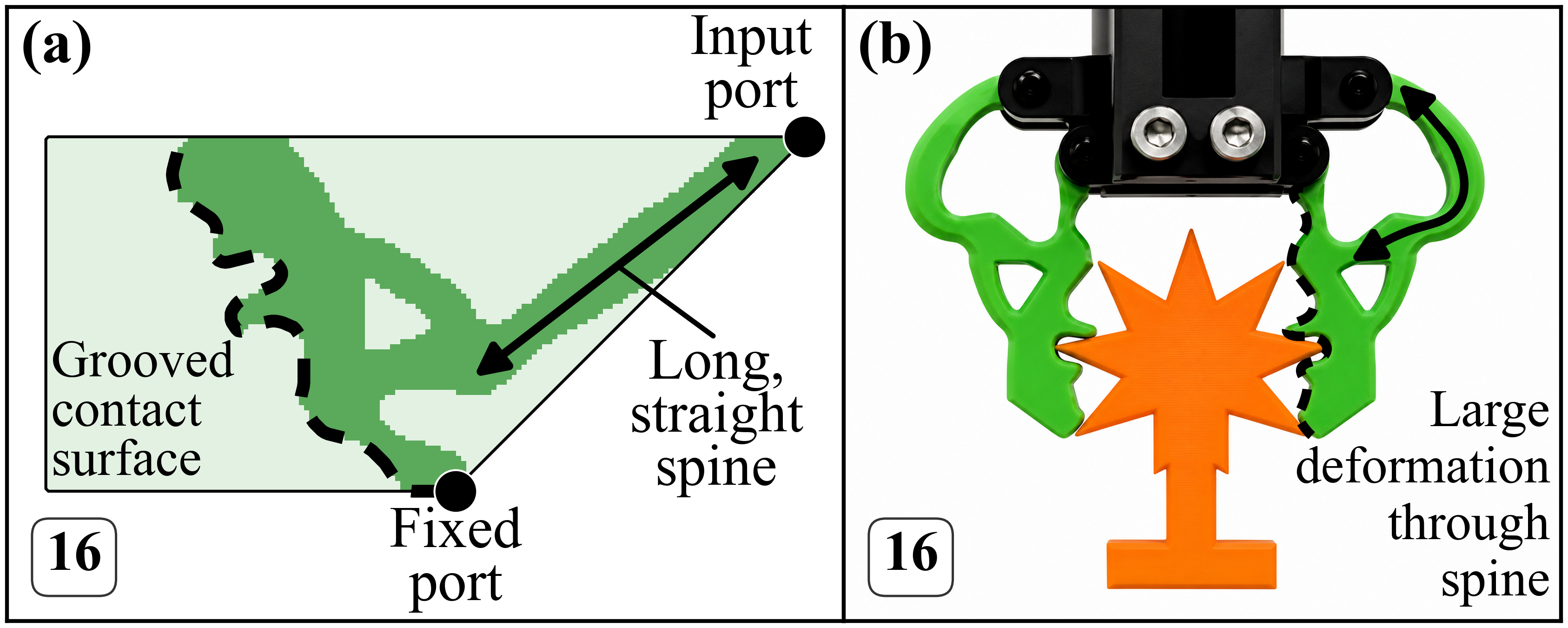}
    \caption{\emph{Our most robust design.} \textbf{(a)} Design 16 from Fig.~\ref{fig:results}. \textbf{(b)} Fabricated design grasping the star.}
\label{fig:robustdesign}
\end{figure}

High-performing designs often had two key features, both illustrated by our most robust design (Fig.~\ref{fig:robustdesign}(a)). First, they often possessed a long, straight spine that runs parallel to the line connecting the input and fixed ports. During actuation, this spine deforms substantially, generating sufficient force to grasp objects securely (Fig.~\ref{fig:robustdesign}(b)). In contrast, designs with shorter and/or more angled spines generally gripped less firmly and performed worse (e.g., designs 18, 23, 32, and 33 in Fig.~\ref{fig:results}). A notable exception was design 28, whose highly angled, curved spine enabled it to wrap securely around objects. High-performing designs also typically possessed large, grooved contact surfaces that conformed to different object geometries without sticking to object features during pick-and-place. Other high-performing designs (14, 17, 25, 29, and 37 in Fig.~\ref{fig:results}) possessed similar high-force-generating spines and broad, adaptable contact surfaces.

Conversely, the two rigid grippers and the best SimTO gear design (design 6 in Fig.~\ref{fig:results}) demonstrated that grasp strength alone does not guarantee robustness. Although these designs held objects securely once grasped, their contact surfaces accommodated little variation in object geometry or grasp pose, leading to frequent outright failures, as illustrated by the rigid gripper at pose $X_1$ in Fig.~\ref{fig:exp_setup}(j). Overall, these observations suggest that robust grasping requires a balance between grasp strength and geometric adaptability. A robust gripper must generate enough force to retain objects under disturbance, while a large, grooved contact surface helps accommodate variations in object geometry and grasp pose.

\section{Conclusion}\label{sec:conclusion}

We have presented SimTO-Robust, a computational design framework that optimises soft gripper morphologies for robustness across different objects and grasp poses. In physical experiments spanning five objects and six grasp poses, SimTO-Robust designs achieved an 86.2\% pick-and-place success rate, compared with 67.3\% for specialised SimTO designs. The best SimTO-Robust designs also withstood more external disturbances than any baseline rigid or soft gripper. Simulation studies further showed that incorporating multiple grasp poses during optimisation increases robustness to grasp height. Grasp robustness, however, peaks at intermediate load-set sizes, beyond which additional simulated objects and poses increase computational cost without improving performance. Varying the simulated grasp pose between optimisation iterations offers a more computationally efficient route to robustness. Overall, these findings demonstrate that grasp robustness can be embedded into a gripper's morphology rather than relying solely on complex sensing and control. Future work will extend SimTO-Robust to 3D topology optimisation, where pose variations in $x$, $y$, and $\theta_z$ are expected to have a greater influence on optimised morphologies.